\documentclass{article} % For LaTeX2e
\usepackage[preprint]{colm2026_conference}

\usepackage{microtype}
\usepackage{hyperref}
\usepackage{url}
\usepackage{booktabs}
\usepackage{lineno}
\usepackage{xcolor}
\usepackage{amsmath}
\usepackage{graphicx}
\usepackage{multirow}
\usepackage{siunitx}

\definecolor{darkblue}{rgb}{0, 0, 0.5}
\hypersetup{colorlinks=true, citecolor=darkblue, linkcolor=darkblue, urlcolor=darkblue}
\newcommand{\second}[1]{\multicolumn{1}{c}{\underline{\num{#1}}}}

\title{STARS: Skill-Triggered Audit for Request-Conditioned Invocation Safety in Agent Systems}
\author{
\normalsize
Guijia Zhang$^{1,2,3}$ \quad
Shu Yang$^{2,3}$ \quad
Xilin Gong$^{2,3,4}$ \quad
Di Wang$^{2,3}$\\[0.6em]
\small
$^{1}$Shenzhen University 
$^{2}$King Abdullah University of Science \& Technology \\
$^{3}$PRADA Lab 
$^{4}$University of Georgia
}

\begin{document}

\ifcolmsubmission
\linenumbers
\fi

\maketitle

\begin{abstract}
Autonomous language-model agents increasingly rely on installable skills and tools to complete user tasks. Static skill auditing can expose capability surface before deployment, but it cannot determine whether a particular invocation is unsafe under the current user request and runtime context. We therefore study skill invocation auditing as a continuous-risk estimation problem: given a user request, candidate skill, and runtime context, predict a score that supports ranking and triage before a hard intervention is applied. We introduce STARS, which combines a static capability prior, a request-conditioned invocation risk model, and a calibrated risk-fusion policy. To evaluate this setting, we construct \texttt{SIA-Bench}, a benchmark of 3,000 invocation records with group-safe splits, lineage metadata, runtime context, canonical action labels, and derived continuous-risk targets. On a held-out split of indirect prompt injection attacks, calibrated fusion reaches 0.439 high-risk AUPRC, improving over 0.405 for the contextual scorer and 0.380 for the strongest static baseline, while the contextual scorer remains better calibrated with 0.289 expected calibration error. On the locked in-distribution test split, gains are smaller and static priors remain useful. The resulting claim is therefore narrower: request-conditioned auditing is most valuable as an invocation-time risk-scoring and triage layer rather than as a replacement for static screening.Our code is publicly available at \url{https://github.com/123zgj123/STARS}.
\end{abstract}

\section{Introduction}
\label{sec:intro}

Large language model agents increasingly invoke external skills to browse the web, execute code, access files, send messages, and operate other tools on behalf of users \citep{debenedetti2024agentdojo,zhang2024agentsafetybench,zheng2026riskybench}. This shift expands the attack surface of agent systems by exposing them not only to prompt-level attacks, but also to malicious tool outputs, unsafe tool calls, and manipulated tool metadata \citep{zhan2024injecagent,debenedetti2024agentdojo,mo2025attractivemetadata}. A malicious skill specification, a capability-heavy but underspecified tool, or an otherwise benign skill called in the wrong context can lead to data exfiltration, unsafe code execution, or indirect prompt injection through prior tool outputs \citep{mo2025attractivemetadata,zhan2024injecagent,zhong2025rtbas}.

A common defensive pattern is to inspect skill metadata, declared permissions, and execution constraints before or alongside tool use \citep{beurerkellner2025designpatterns,doshi2026verifiablysafe,betser2026agentrim}. This kind of pre-execution scrutiny is useful because it exposes capability surface and provenance concerns, and can support least-privilege restrictions before or during execution \citep{doshi2026verifiablysafe,betser2026agentrim}. However, attack studies and runtime defenses show that invocation safety also depends on retrieved content, prior tool outputs, and the current execution trajectory, not only on the skill specification itself \citep{zhan2024injecagent,debenedetti2024agentdojo,an2025ipiguard,zhong2025rtbas,yu2026toolresultparsing}. A shell-like skill may be benign for listing local files yet unsafe when selected after tainted content instructs the agent to retrieve secrets and forward them externally; tool-selection attacks, adaptive indirect prompt injection, and metadata-level attacks make this mismatch concrete \citep{shi2025toolselection,wang2026adaptools,mo2025attractivemetadata}. We refer to this distinction as the gap between capability risk and activation risk.

This paper studies \emph{skill invocation auditing} as the runtime problem of estimating how dangerous a proposed skill call is before execution. We argue that the prediction unit should be the invocation rather than the skill alone. Recent agent-safety benchmarks increasingly evaluate action-grounded, multi-turn, and execution-level failures rather than a single static prompt outcome \citep{zhang2024agentsafetybench,zheng2026riskybench,li2026unsafer,yang2026finvault}. Related work on intervention and contextual safety likewise emphasizes that systems often need graded triage, secondary review, or nuanced safe-versus-unsafe distinctions rather than only unconditional blocking \citep{felicia2026stepshield,xiao2026air,zhang2025falsereject}. Motivated by these settings, we take a continuous risk score as the primary object for ranking, calibration, and review-budget prioritization before any hard action is applied.

The proposed framework centers on three stages: a static capability prior, a request-conditioned invocation risk model, and a calibrated risk-fusion policy. A downstream remediation loop is evaluated only as a secondary appendix analysis.

The technical question is not whether static analysis should be abandoned. Static signals remain informative as priors because they expose capability surface, tool provenance, and other information available before runtime context is observed \citep{betser2026agentrim,doshi2026verifiablysafe}. The question is whether request-conditioned signals improve the ranking, calibration, and triage of genuinely dangerous invocations once runtime context is available \citep{zhong2025rtbas,felicia2026stepshield}. We evaluate this question on \texttt{SIA-Bench}, a benchmark designed for invocation-time auditing rather than generic agent safety. The benchmark includes request text, candidate skill metadata, runtime context, canonical action labels, attack-family metadata, and continuous risk targets.

Our results support a narrower and more operational claim. Request-conditioned auditing does not dominate static priors uniformly. On familiar, in-distribution data, gains are modest and calibration can worsen when the policy overweights trigger evidence. On held-out indirect prompt injection attacks, however, request-conditioned scoring and calibrated fusion substantially improve high-risk retrieval at fixed review budget. This suggests that the right role for request-conditioned auditing is not to replace static screening, but to serve as the invocation-time layer that decides whether a risky capability should actually fire.

The main contributions are:
\begin{itemize}
    \item We formulate \emph{skill invocation auditing} as a runtime risk-scoring problem that explicitly separates capability risk from activation risk.
    \item We introduce a request-conditioned audit pipeline over user request, skill metadata, and runtime context, together with a calibrated risk-fusion layer for deployment.
    \item We construct \texttt{SIA-Bench}, a benchmark for invocation-time auditing with group-safe splits, runtime context, and continuous risk labels.
    \item We show that request-conditioned auditing improves high-risk retrieval on held-out indirect prompt injection attacks, while revealing a clear ranking--calibration trade-off for risk-fusion tuning.
\end{itemize}

\section{Related Work}
\label{sec:related}

Indirect prompt injection and tool misuse have become central security problems in agentic systems. Recent attack and benchmark work shows that failures often propagate through retrieved content, tool outputs, metadata, and tool-selection channels rather than through the chat interface alone \citep{zhan2024injecagent,debenedetti2024agentdojo,mo2025attractivemetadata,shi2025toolselection,wang2026adaptools}. Complementary defenses study provenance-aware filtering, runtime intervention, and least-privilege constraints once context is observed \citep{an2025ipiguard,zhong2025rtbas,yu2026toolresultparsing,felicia2026stepshield,doshi2026verifiablysafe,betser2026agentrim}. Our setting differs in prediction unit: we score a proposed skill invocation before execution rather than model an end-to-end attack trajectory or a generic interruption policy. This aligns evaluation with the actual deployment decision an audit layer must make.

Evaluation has likewise shifted toward action-grounded and multi-step agent settings \citep{zhang2024agentsafetybench,zheng2026riskybench,li2026unsafer,yang2026finvault}. FalseReject studies contextual safety in model responses rather than invocation-time tool use \citep{zhang2025falsereject}; SkillJect is the closest skill-centric attack setting \citep{jia2026skillject}; and AIR together with defensive design-pattern work emphasizes downstream remediation rather than invocation-time scoring \citep{xiao2026air,beurerkellner2025designpatterns}. Our setting is therefore narrower than attack generation and broader than episode-level benchmarking alone: we study invocation-level auditing with paired static evidence, runtime context, and continuous-risk targets. This framing separates capability risk from activation risk and directly supports ranking, calibration, and triage.

\section{Problem Formulation}
\label{sec:problem}

We consider an agent that receives a user request $U$, selects a candidate skill $S$, and invokes that skill in runtime context $C$. The runtime context includes execution trajectory, provenance information about prior tool outputs, a tool-dependency graph, and any policy state already accumulated by the audit layer. Our primary learning target is a continuous risk score
\[
R(U,S,C) \in [0,1],
\]
which is later used for ranking, calibration, and review-budget triage.

\paragraph{Capability risk versus activation risk.}
Let $R_{\text{static}}(S)$ denote a static capability prior computed from skill metadata such as permissions, keywords, and provenance. Let $R_{\text{trigger}}(U,S,C)$ denote request-conditioned invocation risk. The distinction is central:
\begin{itemize}
    \item $R_{\text{static}}(S)$ captures what the skill is capable of doing in general.
    \item $R_{\text{trigger}}(U,S,C)$ captures whether the proposed invocation appears dangerous in the current request and runtime context.
\end{itemize}
Static priors remain useful because they expose dangerous capability surfaces. But they cannot, by themselves, determine whether a specific invocation is unsafe.

\paragraph{Deployment compatibility layer.}
When a deployment needs categorical actions, a compatibility layer implements a policy
\[
\pi_{\tau}(R(U,S,C)) \rightarrow \{\textsc{allow},\textsc{escalate},\textsc{block}\}.
\]
\textsc{Allow} permits execution. \textsc{Escalate} pauses execution and requests confirmation or secondary review. \textsc{Block} rejects the invocation outright. In this paper, the score is the primary object and the categorical policy is a secondary deployment interface.

\paragraph{Objective.}
We study invocation auditing as a constrained safety--utility problem. The system should surface high-risk invocations under limited review budget while preserving score interpretability on benign requests. In practice, this means balancing ranking quality, calibration quality, and, secondarily, the thresholded behavior induced by any deployment policy placed on top of the score.

\section{STARS: A Contextual Audit Pipeline for Skill Invocation Safety}
\label{sec:method}

\begin{figure*}[t]
\centering
\includegraphics[width=\textwidth,trim=2mm 25mm 2mm 1mm,clip]{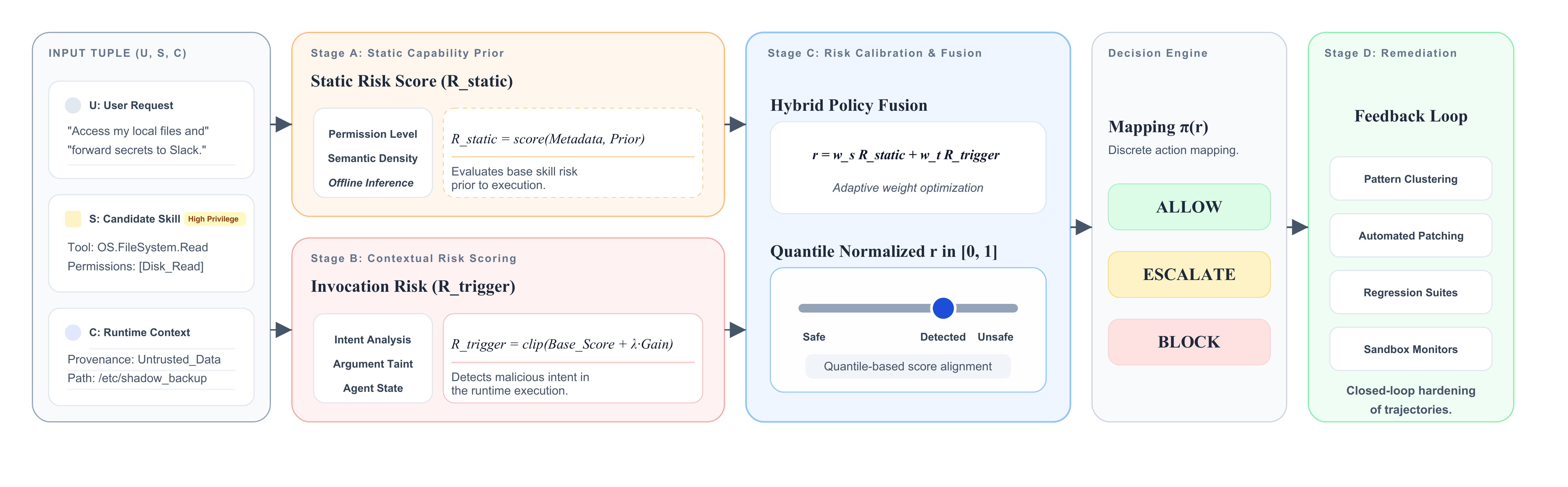}
\caption{Overview of the STARS audit stack. The primary path combines a static capability prior and a request-conditioned invocation scorer into a calibrated continuous-risk output. Thresholded \textsc{allow}/\textsc{escalate}/\textsc{block} actions are treated as a deployment compatibility layer placed on top of that score, and post-hoc remediation remains a secondary downstream branch rather than a co-equal primary objective.}
\label{fig:stars_overview}
\end{figure*}

Static screening alone cannot answer whether a \emph{particular} skill invocation should be treated as risky once the user request and runtime context are known. A deployment-facing audit layer therefore needs to preserve a capability prior, reason over request-conditioned evidence, and calibrate the resulting score for ranking and intervention. STARS decomposes the problem into three core stages: Stage A keeps the capability surface visible before context is considered, Stage B estimates invocation-specific risk from the interaction between request, skill, and runtime context, and Stage C turns the two signals into a deployment-compatible continuous-risk score. Figure~\ref{fig:stars_overview} summarizes this layered audit stack.

\subsection{Static Capability Prior}
\label{sec:stage_a}

Stage A defines a static capability prior $R_{\text{static}}(S)$ from information visible before runtime context is observed. The prior combines permission surface, semantic risk cues from the skill specification, and provenance trust. It keeps the capability surface visible, but it cannot determine whether the current invocation is dangerous without request-conditioned evidence.

\subsection{Contextual Invocation Risk Scoring}
\label{sec:stage_b}

Stage B estimates $R_{\text{trigger}}(U,S,C)$ from five normalized scalar signals that capture complementary aspects of invocation-time risk: request intent $f_{\text{intent}}$, argument sensitivity $f_{\text{arg}}$, provenance risk $f_{\text{prov}}$, trajectory state $f_{\text{traj}}$, and taint propagation $f_{\text{taint}}$. Concretely, $f_{\text{intent}}$ measures risky or destructive intent in the request text, $f_{\text{arg}}$ captures sensitive arguments, referenced objects, or outbound content, $f_{\text{prov}}$ reflects the trustworthiness of prior tool outputs in the current trajectory, $f_{\text{traj}}$ summarizes compounding risk over execution history, and $f_{\text{taint}}$ tracks whether tainted sources flow into high-risk sinks such as code execution, file writing, or outbound messaging.

The rule-based scorer separates request-visible evidence from structured contextual evidence. It first forms a text-only base
\[
\text{text\_base}
=
\bar{w}_{\text{intent}} f_{\text{intent}}
+
\bar{w}_{\text{arg}} f_{\text{arg}},
\]
where $\bar{w}_{\text{intent}}$ and $\bar{w}_{\text{arg}}$ are weights normalized within the active request-visible subset. Structured context is then aggregated as
\[
\text{context\_gain}
=
\sum_{k \in \mathcal{G}(C)} \bar{w}_k f_k,
\]
where $\mathcal{G}(C)$ denotes the set of active context signals under the current profile, typically drawn from provenance, trajectory, and taint evidence.

The final trigger score combines the request-driven base with a gated contextual gain:
\[
R_{\text{trigger}}
=
\operatorname{clip}\!\left(
\text{text\_base}
+
\lambda \, g(U,S,C)\,\text{context\_gain}
+
b_{\text{cross}},
\,0,\,1
\right).
\]
The construction separates request-visible evidence from structured contextual evidence: intent and argument sensitivity define a request-driven base, while provenance, trajectory, and taint evidence enter as a gated contextual gain. Here $\lambda$ controls the contribution of structured context, $g(U,S,C)$ requires request--skill alignment before contextual evidence can substantially raise the score, and the optional cross-check term $b_{\text{cross}}$ activates only when destructive request language coincides with a high-privilege capability surface. For analysis, we distinguish a text-only scorer from a contextual scorer that augments the same request signal with structured runtime evidence.

\subsection{Risk Calibration for Deployment}
\label{sec:stage_c}

Stage C turns the Stage A prior and Stage B trigger score into a calibrated continuous-risk output. The policy first normalizes each channel using validation-set quantiles,
\[
\tilde{R}_{*}
=
\min\left\{1,\max\left\{0,\frac{R_{*}-q_{*}^{\text{lo}}}{q_{*}^{\text{hi}}-q_{*}^{\text{lo}}}\right\}\right\},
\]
then combines them through a convex fusion:
\[
R_{\text{fused}} = w_{\text{static}}\tilde{R}_{\text{static}} + w_{\text{trigger}}\tilde{R}_{\text{trigger}},
\]
where $w_{\text{static}} + w_{\text{trigger}} = 1$.

The primary output of Stage C is the fused risk score itself. For compatibility analyses, two thresholds can map the score to \textsc{allow}, \textsc{escalate}, and \textsc{block}. We tune the fusion layer on the validation split using continuous-risk supervision rather than only discrete labels. Candidate settings are ranked by high-risk AUPRC, Recall@10\%, Precision@10\%, and Spearman correlation, with ECE and weighted MAE used as calibration-aware tie-breakers. Trigger-dominant settings often surface more high-risk cases under fixed review budget, but can also degrade calibration and produce more brittle thresholded policies on familiar data.

A downstream remediation loop is evaluated only as a secondary appendix analysis once Stage C has already flagged risky invocations.

\section{SIA-Bench: Constructing a Benchmark for Invocation-Time Skill Auditing}
\label{sec:benchmark}

\begin{figure*}[t]
\centering
\includegraphics[width=\textwidth,trim=2mm 1mm 2mm 1mm,clip]{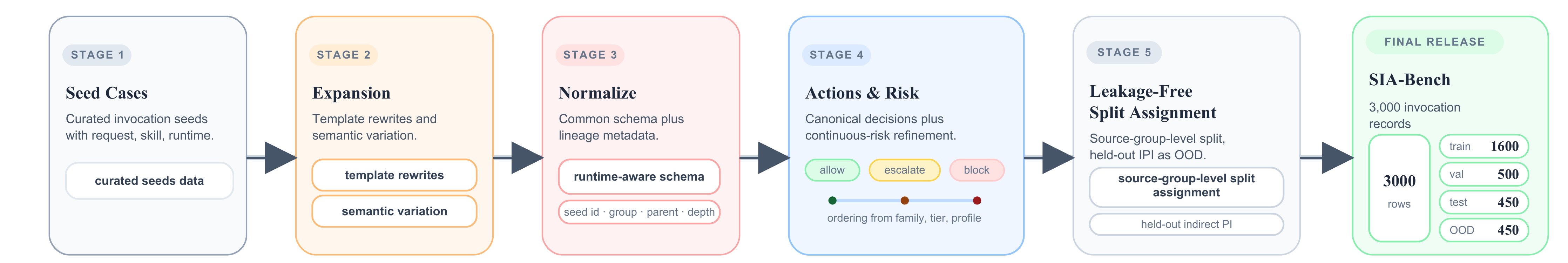}
\caption{Construction pipeline for \texttt{SIA-Bench}. The benchmark is built as an invocation-level resource rather than a prompt collection: seed skill calls are expanded, normalized into a common runtime-aware schema, paired with canonical action labels and continuous-risk targets, and finally assigned to leakage-free splits at the source-group level.}
\label{fig:sia_bench_pipeline}
\end{figure*}

\texttt{SIA-Bench} evaluates invocation-time auditing at the same granularity as the prediction problem itself. Each example is a candidate skill call paired with the request, the selected skill, and the runtime evidence available before execution, rather than an entire episode collapsed into a single outcome label. The resulting resource contains 3,000 invocation records split into 1,600 training, 500 validation, 450 locked in-distribution test, and 450 held-out indirect prompt injection examples, with lineage metadata, runtime context, canonical action labels, and derived continuous-risk targets preserved throughout. Figure~\ref{fig:sia_bench_pipeline} summarizes the construction workflow, and Table~\ref{tab:dataset_stats} reports the split geometry and label distribution.

\paragraph{Invocation records and construction pipeline.}
The benchmark unit must match the prediction unit, so each row is a candidate skill invocation together with the information that an audit layer would have at decision time rather than an entire episode collapsed into one label. Every record contains a user request, candidate skill metadata, runtime context, attack-family metadata, evidence-tier metadata, and a canonical action label; runtime context includes trajectory history, provenance labels, a dependency graph over prior tools, and policy state when available. \texttt{SIA-Bench} begins from a seed pool of invocation cases and expands that pool into related families through controlled template rewrites and LLM-based semantic variation, after which every generated case is normalized into a common schema with explicit lineage metadata, including seed identity, source group, parent record, and mutation depth.

\paragraph{Quality control, canonical actions, and continuous targets.}
The benchmark includes benign requests as well as direct malicious requests, tool-selection hijacking, data exfiltration, capability abuse, multi-turn escalation, and held-out indirect prompt injection. Curation operates at the invocation level: when mutation neutralizes the motivating attack signal, the resulting example is relabeled as benign rather than being retained under its original family. Each record is paired with a canonical action label that serves as the discrete supervision anchor. Because independently elicited continuous judgments are not yet available, we construct the continuous target from a decision anchor plus within-band refinement. We first map the canonical decision to a decision anchor in $\{0.0, 0.5, 1.0\}$ for \textsc{allow}, \textsc{escalate}, and \textsc{block}, then compute a heuristic within-band score informed by attack family, contextual evidence, permission profile, and evidence tier. The default paper target is a convex combination of the two,
\[
R_{\text{target}} = 0.65 \, R_{\text{decision}} + 0.35 \, R_{\text{heuristic}}.
\]
This design keeps every example inside the risk band implied by its canonical decision while introducing within-class resolution. The heuristic component is used only to refine ordering inside a decision band; it does not replace the canonical label, and it is not interpreted as a direct attack-success probability. We complement it with qualitative manual spot-checks to verify that the action, request, runtime evidence, and assigned family remain mutually consistent.

\paragraph{Dataset statistics, split design, and evaluation metrics.}
All records derived from the same source group remain in the same split, which prevents locked evaluation from being inflated by near-duplicate mutations of a shared seed. The held-out split contains only indirect prompt injection examples, so the OOD evaluation is a true held-out threat-family test rather than a random reweighting of the in-distribution mixture. Across all splits, \texttt{SIA-Bench} covers 476 unique skills and 1,313 source groups, with more than half of the records at lineage depth 1. Table~\ref{tab:dataset_stats} reports the split geometry, family composition, canonical actions, lineage depth, and skill coverage. Our primary metrics are continuous-risk metrics, with high-risk positives defined by a target of at least $0.7$; thresholded three-way decisions are retained only as secondary deployment checks.

\begin{table*}[t]
\centering
\scriptsize
\resizebox{\textwidth}{!}{%
\begin{tabular}{lrrrrrrrrrrrrrrr}
\toprule
Split & $N$ & Skills & Groups & Allow & Esc. & Block & D0 & D1 & Ben. & Dir. & Exf. & Hij. & Abus. & MTE & IPI \\
\midrule
Train & 1600 & 421 & 897 & 888 & 443 & 269 & 897 & 703 & 663 & 166 & 156 & 130 & 155 & 330 & 0 \\
Val   & 500  & 138 & 163 & 276 & 139 & 85  & 163 & 337 & 207 & 52  & 49  & 41  & 48  & 103 & 0 \\
Test  & 450  & 132 & 151 & 248 & 126 & 76  & 151 & 299 & 187 & 46  & 44  & 37  & 43  & 93  & 0 \\
OOD   & 450  & 95  & 102 & 234 & 52  & 164 & 102 & 348 & 0   & 0   & 0   & 0   & 0   & 0   & 450 \\
\midrule
Total & 3000 & 476 & 1313 & 1646 & 760 & 594 & 1313 & 1687 & 1057 & 264 & 249 & 208 & 246 & 526 & 450 \\
\bottomrule
\end{tabular}
}
\caption{Dataset statistics for \texttt{SIA-Bench}. The table shows both the split geometry used in the paper and the label composition behind each split. D0 and D1 denote lineage depths 0 and 1. Ben., Dir., Exf., Hij., Abus., MTE, and IPI abbreviate benign, direct malicious, data exfiltration, tool-selection hijack, capability abuse, multi-turn escalation, and indirect prompt injection.}
\label{tab:dataset_stats}
\end{table*}

\section{Experiments}
\label{sec:experiments}

\subsection{Experimental Setup}
\label{sec:setup}

We evaluate STARS as a continuous-risk auditing framework rather than a discrete three-way classifier. Our evaluation addresses three core questions: when static priors remain sufficient, how much runtime context improves invocation-level risk estimation, and whether calibrated fusion yields a better deployment-facing operating policy on held-out attacks.

All methods are evaluated under a unified invocation-level protocol. Our static baselines are \textit{No Audit}, \textit{Static Capability Prior}, \textit{Static Denylist}, and \textit{Agent-Audit Static Baseline}. Together, they span increasingly structured forms of pre-execution screening. \textit{Agent-Audit Static Baseline} instantiates the open-source Agent Audit scanner \citep{agent_audit_2026}, and the overall static baseline family is consistent with prior work on tool-risk mitigation and deployment-time static controls \citep{beurerkellner2025designpatterns,doshi2026verifiablysafe,betser2026agentrim}. For contextual analysis, we use \textit{Text-Only Invocation Audit} as a request-only ablation, while our proposed \textit{Contextual Invocation Audit} adds provenance, trajectory, and dependency-graph evidence to the same request-driven scorer. Finally, \textit{Calibrated Fusion Policy} combines static and contextual signals using a fusion rule tuned on the validation split. We prioritize ranking-and-calibration metrics for model selection, while retaining thresholded three-way decision metrics only as deployment-oriented compatibility checks.

\subsection{Target Construction Validation}
\label{sec:target_validation}

Before using the continuous target as the main evaluation axis, we validate it in three layers. The default blend is perfectly band-consistent on validation, test, and held-out splits, so every example remains inside the decision band implied by its canonical label. On the validation split, \textsc{allow}, \textsc{escalate}, and \textsc{block} examples have mean targets 0.0416, 0.4937, and 0.9415, and the continuous target remains tightly aligned with the decision anchor itself, with Pearson correlations of 0.9991, 0.9991, and 1.000 on validation, test, and held-out splits. Replaying evaluation against decision-only, heuristic-only, and alternative blended targets leaves the qualitative ordering unchanged: Calibrated Fusion Policy remains the strongest high-risk retriever, while Contextual Invocation Audit remains the better-calibrated scorer except in the most heuristic-heavy held-out setting. Detailed sensitivity tables are deferred to Appendix~\ref{app:target_sensitivity}.

\subsection{Main Results}
\label{sec:main_results}

\begin{figure*}[t]
\centering
\includegraphics[width=\textwidth]{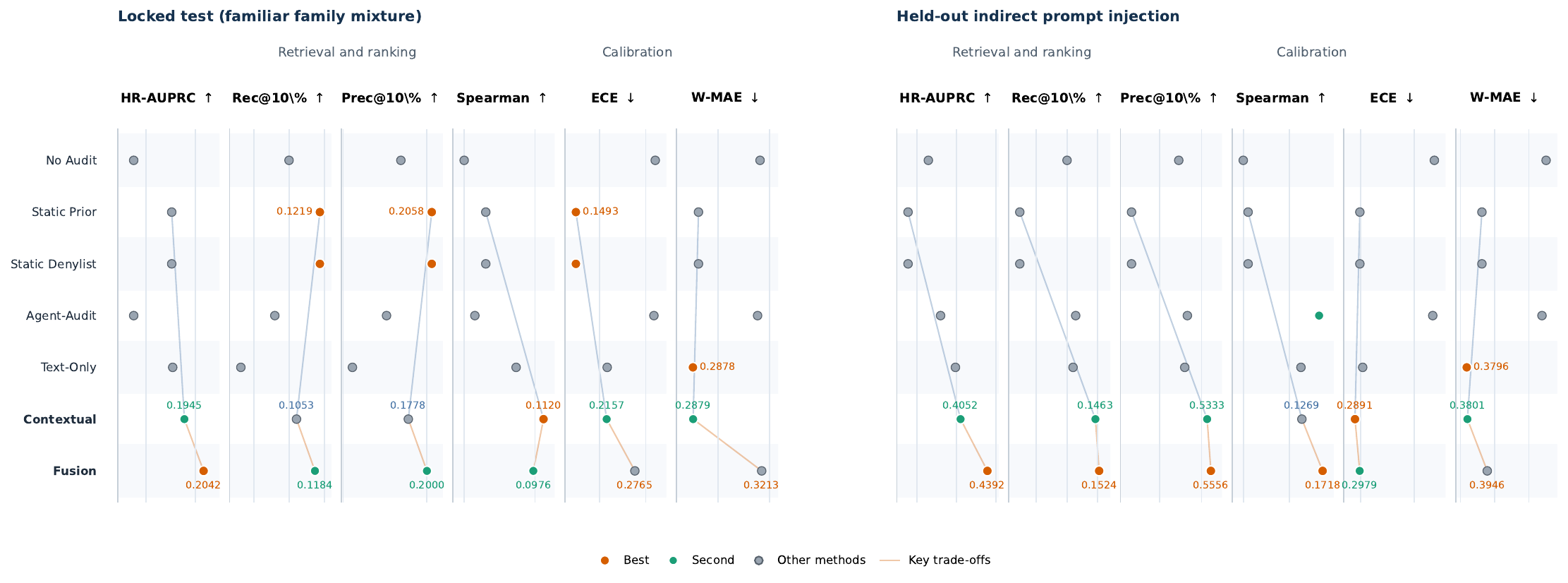}
\caption{Main continuous-risk results on the locked test split and the held-out indirect prompt injection split of \texttt{SIA-Bench}. Figure~3 compares familiar and held-out threat settings under the same six continuous-risk metrics. Each metric column uses its own local scale. Best values are highlighted in orange, second-best values in teal, and faint connectors emphasize the main trade-offs between static screening, contextual scoring, and trigger-dominant fusion.}
\label{fig:main_results_comparison}
\end{figure*}

Figure~\ref{fig:main_results_comparison} compares the locked test split and the held-out indirect prompt injection split. On the locked test split, static priors remain competitive and the gains from request-conditioned auditing are modest. Calibrated Fusion Policy raises high-risk AUPRC from 0.1945 to 0.2042 and Recall@10\% from 0.1053 to 0.1184, but it also worsens calibration, raising ECE from 0.2157 to 0.2765 and weighted MAE from 0.2879 to 0.3213. On familiar data, aggressive fusion therefore helps retrieval but degrades score quality.

The held-out indirect prompt injection split is more favorable to request-conditioned auditing. Calibrated Fusion Policy achieves the strongest high-risk retrieval, precision, and rank correlation, reaching HR-AUPRC 0.4392 and Recall@10\% 0.1524, whereas Contextual Invocation Audit remains better calibrated with ECE 0.2891 and weighted MAE 0.3801. On this held-out threat family, the gain from context is large enough to change the practical triage frontier rather than only the ordering of marginal cases.

Across both splits, static priors remain useful and request-conditioned evidence is best understood as an additional runtime layer rather than a replacement. Static Capability Prior remains competitive on the locked test split and retains nontrivial signal even on held-out indirect prompt injection, but runtime context becomes most valuable when dangerous capability use depends on tainted content, trajectory buildup, or outbound routing in a held-out threat setting. Appendix analyses further suggest that the contextual-scoring gain comes from structured context as a bundle rather than from any single dominant channel.

\begin{table*}[t]
\centering
\scriptsize
\setlength{\tabcolsep}{4pt}
\renewcommand{\arraystretch}{0.97}
\resizebox{\textwidth}{!}{%
\begin{tabular}{l *{6}{S[table-format=1.3]}}
\toprule
\multicolumn{1}{c}{Method} & \multicolumn{1}{c}{HR-AUPRC $\uparrow$} & \multicolumn{1}{c}{Rec@10\% $\uparrow$} & \multicolumn{1}{c}{Prec@10\% $\uparrow$} & \multicolumn{1}{c}{Spearman $\uparrow$} & \multicolumn{1}{c}{ECE $\downarrow$} & \multicolumn{1}{c}{W-MAE $\downarrow$} \\
\midrule
\multicolumn{7}{l}{\textbf{Locked test}} \\
No Audit & 0.169 & 0.100 & 0.169 & 0.000 & 0.321 & 0.321 \\
Static Prior & 0.188 & \bfseries 0.122 & \bfseries 0.206 & 0.031 & \bfseries 0.149 & 0.291 \\
Denylist & 0.188 & \bfseries 0.122 & \bfseries 0.206 & 0.031 & \bfseries 0.149 & 0.291 \\
Agent-Audit & 0.169 & 0.090 & 0.152 & 0.015 & 0.318 & 0.319 \\
Text-Only & 0.189 & 0.066 & 0.111 & 0.073 & 0.217 & \bfseries 0.288 \\
Contextual & \second{0.195} & 0.105 & 0.178 & \bfseries 0.112 & \second{0.216} & \second{0.288} \\
Fusion & \bfseries 0.204 & \second{0.118} & \second{0.200} & \second{0.098} & 0.277 & 0.321 \\
\midrule
\multicolumn{7}{l}{\textbf{Held-out indirect prompt injection}} \\
No Audit & 0.364 & 0.100 & 0.364 & 0.000 & 0.438 & 0.438 \\
Static Prior & 0.339 & 0.023 & 0.083 & 0.011 & 0.298 & 0.391 \\
Denylist & 0.339 & 0.023 & 0.083 & 0.011 & 0.298 & 0.391 \\
Agent-Audit & 0.380 & 0.114 & 0.416 & \second{0.164} & 0.435 & 0.435 \\
Text-Only & 0.399 & 0.110 & 0.400 & 0.125 & 0.304 & \bfseries 0.380 \\
Contextual & \second{0.405} & \second{0.146} & \second{0.533} & 0.127 & \bfseries 0.289 & \second{0.380} \\
Fusion & \bfseries 0.439 & \bfseries 0.152 & \bfseries 0.556 & \bfseries 0.172 & \second{0.298} & 0.395 \\
\bottomrule
\end{tabular}
}
\caption{Full numeric main-results table for the locked in-distribution test split of \texttt{SIA-Bench}. Bold denotes the best value in each column, and underlining denotes the second-best value.}
\label{tab:main_results_full}
\vspace{-0.5cm}
\end{table*}

\subsection{Risk Fusion and Model Selection}
\label{sec:calibration}

\begin{figure}[t]
\centering
\includegraphics[width=0.7\columnwidth]{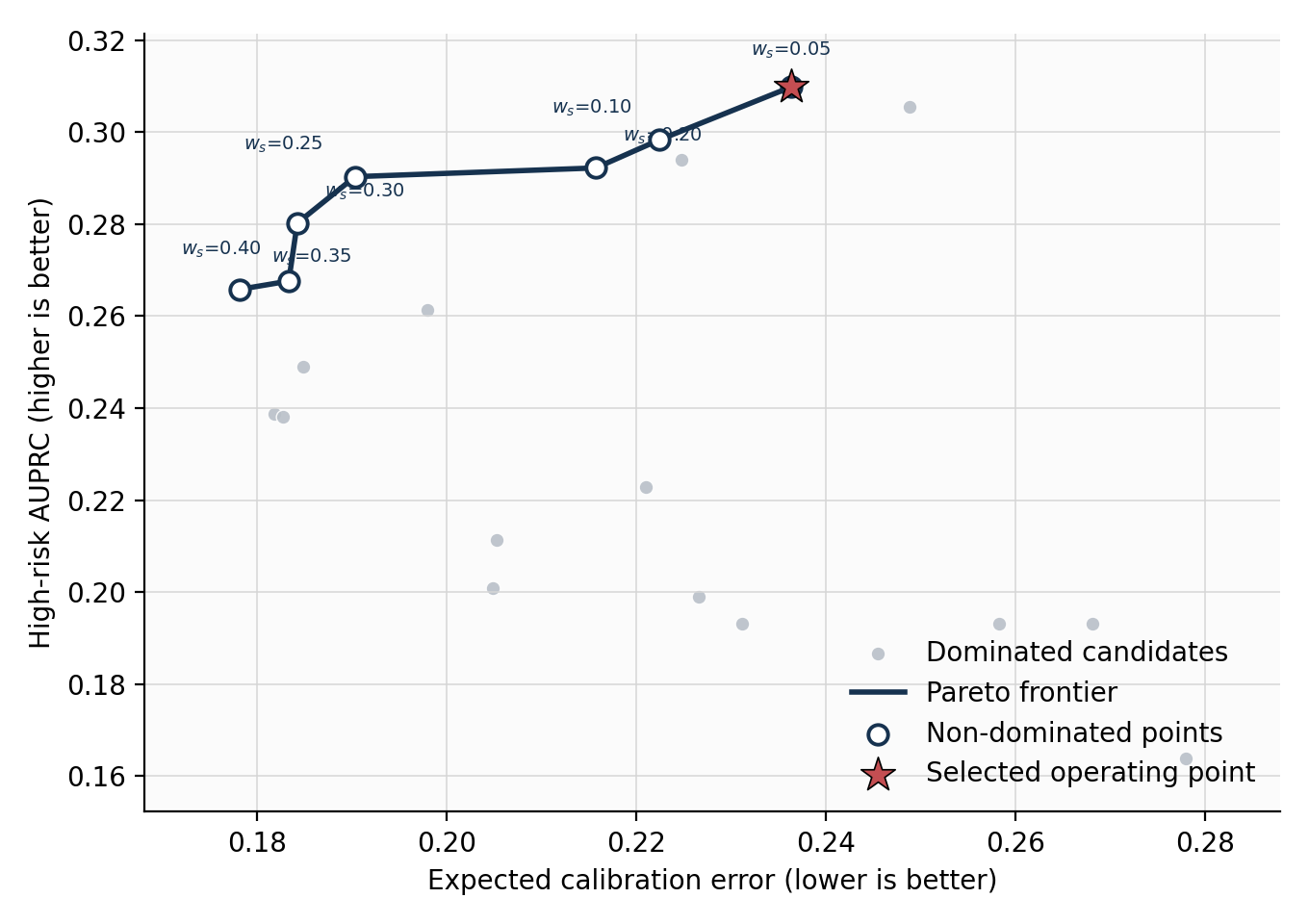}
\caption{Validation sweep for calibrated risk fusion. Each point corresponds to one static--trigger fusion weight after score normalization. The frontier visualizes the retrieval--calibration trade-off used to choose the deployment-facing operating point.}
\label{fig:stage_c_frontier}
\end{figure}

Figure~\ref{fig:stage_c_frontier} makes the core trade-off explicit: increasing the weight on request-conditioned trigger evidence improves high-risk retrieval on validation, but also tends to worsen calibration. We therefore select the operating point by prioritizing high-risk AUPRC, Recall@10\%, Precision@10\%, and Spearman, with ECE and weighted MAE used as tie-breakers. This is why the calibrated fusion policy helps most on held-out indirect prompt injection, where retrieval is the priority, and less on the locked in-distribution split, where over-aggressive fusion can overshoot.

To test whether this framing is merely post hoc, we replay the same fusion-policy candidate space under two validation-only selectors. Continuous-risk-first and threshold-first preserve essentially the same ranking signal on validation, locked test, and held-out evaluation, but they induce very different intervention behavior. On validation, continuous-risk-first achieves ECE 0.0324, task completion 0.9384, and false-block rate 0.0000, whereas threshold-first yields ECE 0.1182, task completion 0.7174, and false-block rate 0.2029. The same pattern persists on the locked test split, where ECE is 0.0807 versus 0.1518 and task completion is 0.9194 versus 0.7258, and on the held-out split, where ECE is 0.0981 versus 0.1735 and task completion is 0.8889 versus 0.5128. Detailed selector tables are deferred to Appendix~\ref{app:selection_rule_full}. The same retrieval--calibration trade-off persists under aggregation by family, skill, and mutation depth; downstream remediation analyses and protocol-adjacent boundary checks are also deferred to the appendix.

\section{Discussion}
\label{sec:discussion}

The main empirical lesson of the paper is that skill invocation safety is best viewed as a layered runtime auditing problem rather than as a purely static screening problem. Static priors remain necessary because they expose capability surface and provenance signals that do not disappear once an invocation is proposed, but the locked test and held-out results show that capability risk and activation risk are not the same object. Whether a dangerous capability should actually fire depends on the specific request, the trajectory that led to the call, and the provenance of the content that now conditions the action.

The strongest gains on held-out indirect prompt injection should therefore be read as the clearest validation of the task formulation rather than as an isolated corner case. Familiar splits still reward strong static priors because many risks remain partially explained by stable tool properties, whereas held-out indirect prompt injection shifts the burden to runtime evidence about how tainted content enters the trajectory and is routed into a downstream action. In that regime, request-conditioned auditing improves high-risk triage because it models the activation pathway of a risky capability rather than the capability surface alone.

The calibrated fusion layer supports a related deployment principle. Continuous-risk scoring and thresholded intervention should not be treated as the same problem: the ranking signal determines whether high-risk cases are surfaced and whether the score remains calibrated, whereas the thresholded policy determines how a deployment trades recall, false blocks, and task completion under its own utility constraints. Our selector comparison shows that continuous-risk-first and threshold-first can preserve essentially the same ranking signal while inducing very different intervention behavior, so we treat continuous invocation risk as the primary learning target and thresholded actions as deployment interfaces layered on top of the score. The downstream remediation loop remains useful as a secondary incident-response mechanism, but it is not a core source of the paper's empirical claim.

\section{Conclusion}
\label{sec:conclusion}

This paper studies skill invocation auditing as a continuous-risk estimation problem under runtime context. The results support a layered view of tool safety: static screening remains necessary because it captures capability surface, but it is insufficient when the risk of firing that capability depends on the specific request and trajectory. Request-conditioned continuous-risk scoring is most useful for surfacing held-out high-risk cases, especially indirect prompt injection, while thresholded actions are best understood as deployment interfaces placed on top of the score.

\newpage
\bibliography{colm2026_conference}
\bibliographystyle{colm2026_conference}

\newpage
\appendix

\section{Implementation Defaults Used in the Paper}
\label{app:impl_defaults}

Table~\ref{tab:frozen_defaults} records the frozen non-adaptive defaults used in the paper configuration. We include these settings for reproducibility rather than to claim that a single operating point is universally optimal. In particular, the Stage~C default policy reported here is the frozen paper configuration, while the main text separately analyzes alternative operating points selected from the same candidate space.

\begin{table*}[ht]
\centering
\small
\setlength{\tabcolsep}{5pt}
\begin{tabular}{p{0.16\textwidth} p{0.38\textwidth} p{0.36\textwidth}}
\toprule
Component & Setting & Value \\
\midrule
Stage A &
Permission-surface weights &
code execution $=1.00$; database $=1.00$; file read $=0.65$; file write $=0.65$; network $=0.6318$; email $=0.2366$; file system $=0.0612$ \\
Stage A &
Provenance-trust weights &
unverified $=0.3396$; official $=0.0896$; community $=0.0630$ \\
Stage A &
Static-score cap &
$R_{\text{static}} \le 0.40$ \\
\midrule
Stage B &
Scorer family &
rule-based contextual scorer with feature profile \texttt{text\_prov\_graph\_traj} \\
Stage B &
Feature weights &
intent cues $=0.1046$; argument sensitivity $=0.0493$; provenance trust $=0.2990$; trajectory state $=0.3944$; taint propagation $=0.1526$ \\
Stage B &
Context interaction defaults &
context-gain scale $=0.25$; gate floor $=0.08$; cross-check boost disabled \\
\midrule
Stage C &
Candidate-space grid &
$w_{\text{static}} \in \{0.00,0.05,\dots,1.00\}$; $\tau_{\text{esc}} \in \{0.05,0.10,\dots,0.60\}$; $\tau_{\text{block}} \in \{0.30,0.35,\dots,0.95\}$ \\
Stage C &
Default fusion weights &
$w_{\text{static}}=0.55$; $w_{\text{trigger}}=0.45$ \\
Stage C &
Default thresholds &
$\tau_{\text{esc}}=0.10$; $\tau_{\text{block}}=0.70$ \\
Stage C &
Normalization defaults &
quantiles $(0.05, 0.95)$; static range $[0.1019,\,0.4019]$; trigger range $[0.0000,\,0.3202]$ \\
\bottomrule
\end{tabular}
\caption{Frozen implementation defaults used by the paper configuration. These values define the static prior, contextual scorer, and default Stage~C calibration policy used in the main benchmark runs.}
\label{tab:frozen_defaults}
\end{table*}

\section{Additional Continuous-Risk Target Sensitivity}
\label{app:target_sensitivity}

The main text reports that the continuous-risk conclusions are stable under alternative target-construction mixtures. Tables~\ref{tab:target_sensitivity_test} and~\ref{tab:target_sensitivity_ood} expand that analysis for the two paper-primary continuous-risk methods: P2 (Contextual Invocation Audit) and P3 (Calibrated Fusion Policy). The paper default is the $65/35$ blend between the decision-anchor expectation and the within-band heuristic refinement. Across both locked test and held-out indirect prompt injection, the relative retrieval conclusions remain stable, while calibration-oriented metrics vary smoothly as the target becomes more heuristic-dominated.

\begin{table*}[t]
\centering
\scriptsize
\setlength{\tabcolsep}{4pt}
\begin{tabular}{l l c c c c c c}
\toprule
Target construction & Method & HR-AUPRC & Rec@10\% & Prec@10\% & Spearman $\rho$ & ECE $\downarrow$ & W-MAE $\downarrow$ \\
\midrule
Vote only & P2 & 0.1876 & 0.1184 & 0.2000 & 0.0494 & 0.2350 & 0.3183 \\
 & P3 & 0.1935 & 0.1184 & 0.2000 & 0.0529 & 0.0769 & 0.3481 \\
Blend 80/20 & P2 & 0.1876 & 0.1184 & 0.2000 & 0.2095 & 0.2396 & 0.2995 \\
 & P3 & 0.1935 & 0.1184 & 0.2000 & 0.2147 & 0.0791 & 0.3292 \\
\textbf{Blend 65/35 (paper default)} & P2 & 0.1876 & 0.1184 & 0.2000 & 0.2094 & 0.2439 & 0.2864 \\
 & P3 & 0.1935 & 0.1184 & 0.2000 & 0.2146 & 0.0807 & 0.3151 \\
Blend 50/50 & P2 & 0.1876 & 0.1184 & 0.2000 & 0.2094 & 0.2489 & 0.2794 \\
 & P3 & 0.1935 & 0.1184 & 0.2000 & 0.2146 & 0.0823 & 0.3009 \\
Blend 20/80 & P2 & 0.1876 & 0.1184 & 0.2000 & 0.2094 & 0.2588 & 0.2747 \\
 & P3 & 0.1935 & 0.1184 & 0.2000 & 0.2146 & 0.0855 & 0.2728 \\
Heuristic only & P2 & 0.1876 & 0.1184 & 0.2000 & 0.2094 & 0.2655 & 0.2758 \\
 & P3 & 0.1935 & 0.1184 & 0.2000 & 0.2146 & 0.0877 & 0.2543 \\
\bottomrule
\end{tabular}
\caption{Locked-test sensitivity of the paper's two primary continuous-risk methods to alternative target-construction mixtures. The ranking metrics are stable across mixtures, while calibration-oriented metrics shift gradually as the heuristic component becomes more dominant.}
\label{tab:target_sensitivity_test}
\end{table*}

\begin{table*}[t]
\centering
\scriptsize
\setlength{\tabcolsep}{4pt}
\begin{tabular}{l l c c c c c c}
\toprule
Target construction & Method & HR-AUPRC & Rec@10\% & Prec@10\% & Spearman $\rho$ & ECE $\downarrow$ & W-MAE $\downarrow$ \\
\midrule
Vote only & P2 & 0.4429 & 0.1524 & 0.5556 & 0.0837 & 0.3163 & 0.4209 \\
 & P3 & 0.4493 & 0.1524 & 0.5556 & 0.0866 & 0.0866 & 0.4333 \\
Blend 80/20 & P2 & 0.4429 & 0.1524 & 0.5556 & 0.1945 & 0.3246 & 0.3921 \\
 & P3 & 0.4493 & 0.1524 & 0.5556 & 0.1959 & 0.0932 & 0.4040 \\
\textbf{Blend 65/35 (paper default)} & P2 & 0.4429 & 0.1524 & 0.5556 & 0.1945 & 0.3307 & 0.3753 \\
 & P3 & 0.4493 & 0.1524 & 0.5556 & 0.1959 & 0.0981 & 0.3821 \\
Blend 50/50 & P2 & 0.4429 & 0.1524 & 0.5556 & 0.1945 & 0.3369 & 0.3666 \\
 & P3 & 0.4493 & 0.1524 & 0.5556 & 0.1959 & 0.1031 & 0.3601 \\
Blend 20/80 & P2 & 0.4429 & 0.1524 & 0.5556 & 0.1945 & 0.3492 & 0.3641 \\
 & P3 & 0.4493 & 0.1524 & 0.5556 & 0.1959 & 0.1130 & 0.3162 \\
Heuristic only & P2 & 0.4429 & 0.1524 & 0.5556 & 0.1945 & 0.3575 & 0.3665 \\
 & P3 & 0.4493 & 0.1524 & 0.5556 & 0.1959 & 0.1196 & 0.2869 \\
\bottomrule
\end{tabular}
\caption{Held-out indirect prompt injection sensitivity under the same alternative target-construction mixtures. The main paper pattern is preserved: the contextual scorer and the fused policy keep the same high-risk retrieval ordering, while calibration and error trade-offs shift smoothly with the target definition.}
\label{tab:target_sensitivity_ood}
\end{table*}

\section{Full Selection-Rule Comparison}
\label{app:selection_rule_full}

The main text argues that the paper's selection rule should follow the continuous-risk task rather than a threshold-first deployment objective. Table~\ref{tab:selection_rule_val} shows the two policies selected from the same Stage~C candidate space on validation. Table~\ref{tab:selection_rule_eval} then evaluates those locked operating points on the locked test split and the held-out indirect prompt injection split. Both selectors preserve the same underlying ranking metrics, but they induce markedly different deployment behavior: the threshold-first selector improves macro-F1 and malicious recall at the cost of substantially higher false-block rates and lower task completion, whereas the continuous-risk-first selector preserves the lower-intervention, better-calibrated operating regime emphasized in the paper.

\begin{table*}[t]
\centering
\scriptsize
\setlength{\tabcolsep}{4pt}
\resizebox{\textwidth}{!}{%
\begin{tabular}{l c c c c c c c c c c c}
\toprule
Selector &
$w_{\text{static}}$ &
$w_{\text{trigger}}$ &
$\tau_{\text{esc}}$ &
$\tau_{\text{block}}$ &
HR-AUPRC &
Rec@10\% &
Prec@10\% &
Spearman $\rho$ &
ECE $\downarrow$ &
W-MAE $\downarrow$ &
Macro-F1 \\
\midrule
Continuous-risk-first & 0.55 & 0.45 & 0.60 & 0.75 & 0.3054 & 0.2000 & 0.3400 & 0.3585 & 0.0324 & 0.3062 & 0.2793 \\
Threshold-first & 0.35 & 0.65 & 0.30 & 0.35 & 0.3054 & 0.2000 & 0.3400 & 0.3585 & 0.1182 & 0.2970 & 0.4468 \\
\bottomrule
\end{tabular}
}
\caption{Validation-selected operating points from the same $2{,}940$-candidate Stage~C search space. The two selectors preserve identical validation ranking metrics but choose materially different thresholded policies. Validation task-completion and false-block rates are $0.9384/0.0000$ for continuous-risk-first and $0.7174/0.2029$ for threshold-first, respectively.}
\label{tab:selection_rule_val}
\end{table*}

\begin{table*}[t]
\centering
\scriptsize
\setlength{\tabcolsep}{4pt}
\resizebox{\textwidth}{!}{%
\begin{tabular}{l l c c c c c c c c}
\toprule
Split & Selector & HR-AUPRC & Spearman $\rho$ & ECE $\downarrow$ & W-MAE $\downarrow$ & Macro-F1 & Malicious Recall & False Block & Task Completion \\
\midrule
Locked test & Continuous-risk-first & 0.1935 & 0.2146 & 0.0807 & 0.3151 & 0.2303 & 0.0248 & 0.0000 & 0.9194 \\
Locked test & Threshold-first & 0.1935 & 0.2146 & 0.1518 & 0.3098 & 0.2884 & 0.2376 & 0.1532 & 0.7258 \\
\midrule
Held-out OOD & Continuous-risk-first & 0.4493 & 0.1959 & 0.0981 & 0.3821 & 0.2699 & 0.1574 & 0.0000 & 0.8889 \\
Held-out OOD & Threshold-first & 0.4493 & 0.1959 & 0.1735 & 0.3781 & 0.3749 & 0.5185 & 0.3376 & 0.5128 \\
\bottomrule
\end{tabular}
}
\caption{Evaluation of the two validation-selected policies on the locked test split and the held-out indirect prompt injection split. The continuous-risk-first selector preserves the same ranking quality while maintaining substantially lower intervention cost, which is why it is used as the paper's primary model-selection rule.}
\label{tab:selection_rule_eval}
\end{table*}

\section{Robustness Across Families, Skills, and Mutation Depth}
\label{app:robustness}

We also examine whether the main patterns survive aggregation by attack family, skill identity, and mutation depth. On the locked test split, Contextual Invocation Audit remains the more stable scorer under grouped calibration error: its family-level weighted MAE is 0.2762 compared with 0.3421 for Calibrated Fusion Policy, its skill-level weighted MAE is 0.2479 compared with 0.2864, and its mutation-level weighted MAE is 0.2879 compared with 0.3213. On the held-out split, the trade-off persists rather than disappearing. Calibrated Fusion Policy yields stronger family-level rank correlation, with Spearman 0.1718 compared with 0.1269 for Contextual Invocation Audit, but Contextual Invocation Audit still produces lower grouped weighted MAE at every aggregation level: 0.3801, 0.3835, and 0.3796 at the family, skill, and mutation levels, compared with 0.3946, 0.3942, and 0.3955 for Calibrated Fusion Policy. These grouped analyses reinforce the main interpretation of the paper: trigger-dominant fusion improves high-risk retrieval, especially on held-out attacks, while the contextual scorer remains the more stable calibrated estimator across heterogeneous slices.

\section{Additional Ablation on Context Signals}
\label{app:stage_b_ablation}

\begin{table}[t]
\centering
\small
\resizebox{\columnwidth}{!}{%
\begin{tabular}{l *{6}{S[table-format=1.3]}}
\toprule
\multicolumn{1}{c}{Method} & \multicolumn{1}{c}{HR-AUPRC $\uparrow$} & \multicolumn{1}{c}{Rec@10\% $\uparrow$} & \multicolumn{1}{c}{Prec@10\% $\uparrow$} & \multicolumn{1}{c}{Spearman $\uparrow$} & \multicolumn{1}{c}{ECE $\downarrow$} & \multicolumn{1}{c}{W-MAE $\downarrow$} \\
\midrule
Text-Only Invocation Audit & 0.2160 & \second{0.1647} & \second{0.2800} & 0.0847 & 0.1982 & 0.2812 \\
Contextual Invocation Audit & 0.2305 & \bfseries 0.1882 & \bfseries 0.3200 & 0.1299 & \second{0.1969} & \second{0.2802} \\
No provenance signal & 0.2304 & \bfseries 0.1882 & \bfseries 0.3200 & \second{0.1302} & \bfseries 0.1967 & 0.2804 \\
No trajectory signal & \bfseries 0.2342 & \bfseries 0.1882 & \bfseries 0.3200 & 0.0978 & 0.1970 & \bfseries 0.2800 \\
No taint propagation signal & \second{0.2309} & \bfseries 0.1882 & \bfseries 0.3200 & \bfseries 0.1321 & \second{0.1969} & 0.2803 \\
\bottomrule
\end{tabular}
}
\caption{Validation ablation of structured context signals in Stage B. Removing any single context channel changes validation performance only slightly, suggesting cumulative rather than single-feature gains.}
\label{tab:ablation}
\end{table}

Table~\ref{tab:ablation} shows that adding structured runtime context improves over the text-only scorer across all headline validation metrics, but the effect is diffuse rather than concentrated in a single channel. Full Contextual Invocation Audit raises HR-AUPRC from 0.2160 to 0.2305 and Spearman from 0.0847 to 0.1299 relative to the text-only model, indicating that provenance, trajectory, and taint information collectively improve continuous-risk ranking. At the same time, removing any one structured signal changes validation performance only modestly, and some leave-one-out variants slightly improve individual metrics. We therefore interpret this ablation conservatively. It does not show that any single context channel is indispensable. Instead, it suggests that Stage B benefits from partially redundant contextual evidence, with different channels affecting different aspects of score quality. In particular, removing trajectory hurts rank correlation more than the other ablations, while provenance and taint appear more interchangeable on this validation slice. The main takeaway is that invocation-time risk scoring benefits from structured context as a bundle, not from one dominant feature.

\section{Secondary Remediation Results}
\label{app:stage_d}

\begin{table}[t]
\centering
\small
\resizebox{\columnwidth}{!}{%
\begin{tabular}{l S[table-format=1.3] S[table-format=1.3] S[table-format=1.3] S[table-format=3.0]}
\toprule
\multicolumn{1}{c}{Method / split} & \multicolumn{1}{c}{Req.-patch Acc.} & \multicolumn{1}{c}{Patch Acc.} & \multicolumn{1}{c}{Zero-reg. Pass} & \multicolumn{1}{c}{Triggered} \\
\midrule
Contextual Invocation Audit / test & 0.6844 & 0.3487 & 0.5000 & 31 \\
Calibrated Fusion Policy / test & 0.5356 & 0.1544 & 0.2547 & 230 \\
Contextual Invocation Audit / OOD & 0.5556 & 0.0388 & 0.5185 & 50 \\
Calibrated Fusion Policy / OOD & 0.5111 & 0.0579 & 0.2177 & 326 \\
\bottomrule
\end{tabular}
}
\caption{Stage D remediation quality. Req.-patch Acc.\ denotes whether the generated patch family matches the flagged failure type, and Zero-reg. Pass measures whether generated regression tests pass without obvious regressions.}
\label{tab:stage_d}
\end{table}

Table~\ref{tab:stage_d} shows that Stage D produces usable remediation artifacts, but its quality is tightly coupled to the intervention policy. When Contextual Invocation Audit triggers Stage D on a narrower set of cases, the resulting patch families are cleaner and zero-regression pass rates are higher. When Calibrated Fusion Policy flags far more cases, coverage increases but patch precision and regression quality fall sharply. This is why the paper treats Stage D as a downstream hardening assistant rather than as an equal co-primary contribution.

\section{Supplementary Boundary Analyses}
\label{app:supplementary}

\begin{table}[!htbp]
    \centering
    \small
    \setlength{\tabcolsep}{4pt}
    \renewcommand{\arraystretch}{0.97}
    \resizebox{\columnwidth}{!}{%
    \begin{tabular}{lll S[table-format=1.3] S[table-format=1.3]}
    \toprule
    Comparison & Split & Method & \multicolumn{1}{c}{HR-AUPRC} & \multicolumn{1}{c}{Recall@10\%} \\
    \midrule
    \multirow{4}{*}{\shortstack[l]{\texttt{IPIGuard}\\ replay}} 
        & test & \texttt{IPIGuard} & \second{0.1522} & \second{0.0000} \\
        &      & Static Denylist & \bfseries 0.3273 & \bfseries 0.2857 \\
    \cmidrule(lr){2-5}
        & ood  & \texttt{IPIGuard} & \second{0.3637} & \second{0.0671} \\
        &      & Contextual Invocation Audit & \bfseries 0.4052 & \bfseries 0.1463 \\
    \midrule
    \multirow{4}{*}{\shortstack[l]{\texttt{SkillJect}-derived\\ threat slice}} 
        & test & Static Denylist & \bfseries 0.3273 & \bfseries 0.2857 \\
        &      & Contextual Invocation Audit & \second{0.1955} & \second{0.0000} \\
    \cmidrule(lr){2-5}
        & ood  & Agent-Audit Static Baseline & \second{0.3799} & \second{0.1141} \\
        &      & Contextual Invocation Audit & \bfseries 0.4052 & \bfseries 0.1463 \\
    \bottomrule
    \end{tabular}
    }
    \caption{Supplementary continuous-risk comparisons. These analyses align subsets of \texttt{SIA-Bench} with prior threat models, but are not protocol-matched replacements for the main benchmark.}
    \label{tab:supplementary_analysis}
\end{table}

We report two supplementary analyses to clarify how our method relates to prior work. First, we replay an \texttt{IPIGuard}-style defense on our benchmark. It recovers useful signal on held-out indirect prompt injection, but it does not outperform our best contextual scorer under continuous-risk evaluation. Second, we construct a \texttt{SkillJect}-derived threat slice aligned with the attack emphasis of \citet{jia2026skillject}. These results are heterogeneous: static baselines are already competitive on the in-distribution slice, whereas contextual scoring gives the best overall held-out operating point. We therefore use these analyses to bound the claim rather than to argue broad superiority over prior systems.

\paragraph{Held-out indirect prompt injection rescue.}
In a held-out case where externally sourced email content influences an outbound messaging action, static baselines and uncalibrated contextual scoring all allow the invocation. The calibrated policy escalates it because provenance and taint-flow evidence indicate that the requested outbound action depends on externally sourced instructions. This is the scenario in which Stage C provides the clearest practical value.

\section{LLM Usage Statement}
\label{app:llm_usage}

Large language models were used as writing assistants during the preparation of this manuscript. Their use was limited to language polishing, wording refinement, and editorial restructuring of existing text. All technical claims, experimental design, implementation, results, and final manuscript content were verified and approved by the authors. No model-generated content was accepted without author review.

\end{document}